\documentclass[conference]{IEEEtran}
\usepackage[utf8]{inputenc}
\usepackage[T1]{fontenc}
\usepackage{cite}
\usepackage{url}
\usepackage{amsmath,amssymb}
\usepackage{hyperref}
\usepackage{graphicx}
\usepackage{booktabs}
\usepackage{array}
\usepackage{tikz}
\usetikzlibrary{arrows.meta,positioning,calc}
\let\oldthebibliography\thebibliography
\renewcommand{\thebibliography}[1]{\oldthebibliography{#1}\setlength{\itemsep}{0pt}\setlength{\parskip}{0pt}}
\title{Autonomous UAV Route Planning for Coverage Maximization in Environmental Monitoring: A Systematic Literature Review}
\author{
\IEEEauthorblockN{Sebastian Jouannet Contreras}
\IEEEauthorblockA{\textit{Master Program in Computer Science} \\
\textit{Universidad del B\'{\i}o-B\'{\i}o}\\
Chill\'an, Chile \\
sebastian.jouannet2101@alumnos.ubiobio.cl}
\and
\IEEEauthorblockN{Carola Figueroa-Flores}
\IEEEauthorblockA{\textit{Dept. of Computer Science and Information Technology} \\
\textit{Universidad del B\'{\i}o-B\'{\i}o}\\
Chill\'an, Chile \\
cfigueroa@ubiobio.cl}}
\begin{document}
\maketitle
\begin{abstract}
Environmental monitoring with unmanned aerial vehicles (UAVs) requires route planning methods that maximize covered area while handling energy limits, operational constraints, and geometric complexity. This paper reports the protocol and preliminary results of an ongoing systematic literature review (SLR) on autonomous UAV route planning for coverage-oriented environmental monitoring. The review follows the PRISMA 2020 framework and searches Scopus and Web of Science for studies published between 2015 and 2026. The protocol focuses on path planning, coverage path planning, and informative path planning, with emphasis on algorithmic families, coverage and energy metrics, obstacle handling, geometric environment representations, and environmental constraints. At the current stage, 562 records have been identified, 161 duplicates have been removed, and 401 unique records have been screened by title, abstract, and keywords. From these, 247 studies were retained for full-text eligibility assessment (235 eligible and 12 borderline records to be resolved during full-text review). A preliminary analysis of the retained studies suggests strong concentration on coverage-oriented formulations, multi-UAV coordination, and energy-aware optimization, while fewer studies explicitly address weather, uncertainty, or obstacle-rich environments. Most retained studies rely on simulation-based validation, highlighting a potential simulation-to-reality gap, and recent publications show increasing interest in reinforcement learning, hybrid optimization, and geometry-aware planning. These early findings indicate an active but fragmented research landscape and support the need for a structured synthesis to identify mature techniques and unresolved gaps for realistic environmental monitoring missions.
\end{abstract}
\section{Introduction}
Unmanned aerial vehicles (UAVs) have become attractive platforms for environmental monitoring because they can rapidly inspect large, irregular, or hard-to-access areas while collecting high-resolution spatial data. However, coverage-oriented mission planning is not a simple waypoint sequencing problem: it must balance monitored area, flight time, energy availability, path feasibility, and environmental complexity. Representative studies already show the breadth of the field, from energy-constrained area coverage and informative terrain monitoring to multi-UAV assignment and geometry-aware planning for complex scenes \cite{energyConstrainedCoverage2021,terrainEnergyCoverage2025,forestConcaveCPP2025,sweepCoverageMultiUAV2020,informativeTerrainIPP2020}.

The literature is growing quickly and spans exact, heuristic, metaheuristic, geometry-based, and learning-based approaches. Recent works also illustrate the diversity of current directions, including terrain-aware metaheuristic optimization, lightweight multi-agent coordination, genetic coverage over disconnected regions, and continuous trajectory optimization \cite{terrain3DMetaheuristic2025,marlLightweightCoverage2026,gaDisconnectedPolygons2024,continuousNonConvexIPP2026}. This variety is promising, but it also makes it difficult to compare assumptions, metrics, and validation practices across studies.

This work-in-progress paper reports the protocol and first screening results of a systematic literature review intended to answer three questions: \emph{RQ1)} which algorithmic families are most used for autonomous UAV route planning with coverage objectives; \emph{RQ2)} which metrics are employed to evaluate the trade-off between coverage and energy cost; and \emph{RQ3)} which technical and environmental limitations remain insufficiently addressed. The contribution of this paper is twofold: a reproducible review protocol tailored to this problem and a preliminary characterization of the retained studies after title--abstract screening.
\section{Review Design}
\subsection{Protocol}
The review follows PRISMA 2020 to structure identification, screening, eligibility, and inclusion stages \cite{prisma2021}. The unit of analysis is any primary study that proposes, evaluates, or compares a UAV route-planning method for coverage, monitoring, or informative exploration under spatial or operational constraints.

The protocol was designed using a PICOC perspective (Population, Intervention, Comparison, Outcome, and Context). The \emph{population} comprises UAVs, drones, or unmanned aerial systems (UAS) applied to environmental monitoring and area coverage. The \emph{intervention} includes path planning, coverage path planning, informative path planning, optimization methods, computational-geometry-based modeling, and obstacle handling. The \emph{comparison} dimension considers alternative algorithmic families, as well as variants with and without energy or spatial restrictions. The main \emph{outcomes} are coverage, useful monitored area, traveled distance, energy consumption, flight time, computational cost, robustness, and geometric feasibility. The \emph{context} is autonomous environmental monitoring over realistic outdoor environments. Table~\ref{tab:rqs} summarizes the review questions and the evidence planned for each.
\begin{table}[t]
\caption{Review questions and planned evidence}
\label{tab:rqs}
\centering\small
\begin{tabular}{p{0.15\columnwidth}p{0.72\columnwidth}}
\toprule
RQ & Planned evidence to extract \\
\midrule
RQ1 & Algorithmic family, optimization strategy, environment representation, and coordination model used for coverage-oriented planning. \\
RQ2 & Reported evaluation metrics related to coverage, distance, energy, runtime, robustness, and mission feasibility. \\
RQ3 & Explicitly modeled limitations, including obstacles, battery restrictions, weather effects, dynamics, sensing assumptions, and unresolved challenges. \\
\bottomrule
\end{tabular}
\end{table}
\subsection{Search Strategy and Eligibility}
The search was conducted in \textbf{Scopus} and \textbf{Web of Science} (Core Collection) over \textbf{2015--2026}, restricted to English-language journal articles and indexed conference papers. These two databases were selected for their broad indexing of computer-science, engineering, and robotics venues, which substantially overlaps the primary content of IEEE Xplore and the ACM Digital Library; extending the search to those interfaces is left as planned future work (Section~\ref{sec:threats}). The query combined three concept groups: (i) UAV-related descriptors; (ii) path-, route-, trajectory-, coverage-, and informative-path-planning concepts; and (iii) coverage-maximization or environmental-monitoring terms. A fourth block constraining energy, obstacle, optimization, or computational-geometry terms was piloted but discarded after a sensitivity check, because it removed relevant coverage-path-planning studies that mention such constraints only in the full text; these dimensions are instead captured during extraction (Table~\ref{tab:extraction}). The exact platform-adapted queries are reported in Fig.~\ref{fig:queries} to support an independently reproducible protocol.
\begin{figure}[t]
\footnotesize
\textbf{Scopus} (Advanced search):
\begin{verbatim}
TITLE-ABS-KEY( ("UAV" OR "unmanned aerial vehicle"
   OR drone OR UAS)
  AND ("path planning" OR "route planning"
   OR "trajectory planning"
   OR "coverage path planning" OR CPP
   OR "informative path planning")
  AND ("coverage maximization" OR "area coverage"
   OR "environmental monitoring" 
   OR "maximum coverage") )
\end{verbatim}
\textbf{Web of Science} (Core Collection): the same three concept blocks were issued in the \texttt{TS=} (topic) field, with Timespan 2015--2026.
\caption{Exact platform-adapted search queries. The Web of Science query reuses the same Boolean concept blocks in the \texttt{TS=} field. A fourth constraint block (energy/obstacle/optimization/computational-geometry) was piloted and removed after a sensitivity check; those dimensions are captured at the extraction stage instead.}
\label{fig:queries}
\end{figure}
Studies were included when they addressed UAV route planning or coverage, reported at least one planning or optimization method, and described explicit performance metrics or an evaluation setting (criteria I1--I5), and were published in English within 2015--2026 (criterion I6). Studies focused exclusively on low-level control, communications, or hardware (E1), using the UAV as a mere capture platform (E2), where coverage is not an objective, constraint, or evaluation metric (E3), on non-UAV platforms without clear methodological transfer (E4), or that were conceptual works/surveys without their own evaluation (E5), were excluded. Duplicate records (E6) were removed before screening. Importantly, the word ``coverage'' is interpreted by \emph{objective function} rather than by keyword: studies optimizing communication coverage, signal range, or number of connected users are excluded (E3), whereas studies optimizing observed area or spatial information are retained. The current paper reports the process up to title--abstract--keyword screening; full-text eligibility and quality assessment are still in progress.
\subsection{Planned Quality Assessment and Synthesis}
The next stages of the review will apply a structured methodological quality checklist to each full-text study. The checklist evaluates objective clarity, algorithm description, environment specification, explicit operational constraints, reported metrics, comparison against established baselines (e.g., classic genetic algorithms, standard lawnmower patterns, or A$^\ast$ variants) rather than only self-generated variants, computational-efficiency reporting, and discussion of limitations. Each item will be scored on a 0, 0.5, or 1 scale. In parallel, a structured extraction form will record algorithmic family, environment representation, coverage model, sensor assumptions, constraints, reported metrics, validation scenario, and comparison baselines. Following common terminology, the extraction explicitly separates \emph{heuristics} (problem-specific constructive rules, e.g., lawnmower/boustrophedon, greedy, spiral) from \emph{metaheuristics} (general-purpose search, e.g., genetic algorithms, ACO, PSO), since conflating them obscures runtime-versus-optimality trade-offs. Table~\ref{tab:extraction} lists the planned extraction dimensions.
\begin{table}[t]
\caption{Planned extraction dimensions for full-text analysis}
\label{tab:extraction}
\centering\small
\begin{tabular}{p{0.34\columnwidth}p{0.56\columnwidth}}
\toprule
Dimension & Examples of extracted information \\
\midrule
Method family & Exact, heuristic (constructive), metaheuristic, learning-based, hybrid \\
Environment representation & Grid, graph, polygon, raster, voxel, continuous map \\
Operational constraints & Energy, time, altitude, obstacles, no-fly zones, weather \\
Evaluation outputs & Coverage, distance, energy, runtime, robustness, success rate \\
Baseline comparison & Established benchmark, self-generated variant, none \\
Validation setting & Simulation, field test, real data, mixed evaluation \\
\bottomrule
\end{tabular}
\end{table}
The final synthesis is expected to combine descriptive, thematic, and comparative analyses, allowing the review to contrast algorithmic choices with the type of environment representation, operational realism, and evaluation practice used in the literature.
\section{Preliminary Screening Results}
\begin{figure}[t]
\centering
\begin{tikzpicture}[
  font=\scriptsize,
  box/.style={draw, rounded corners, align=center, inner sep=3pt, text width=4.25cm},
  sidebox/.style={draw, dashed, rounded corners, align=left, inner sep=3pt, text width=2.55cm},
  arrow/.style={-{Latex[length=1.6mm]}, thick}
]
\node[box] (id) {Records identified from Scopus and Web of Science\\\textbf{n = 562} (Scopus 384; WoS 178)};
\node[box, below=0.35cm of id] (dup) {Duplicates removed\\\textbf{n = 161}};
\node[box, below=0.35cm of dup] (screen) {Unique records screened by title, abstract, and keywords\\\textbf{n = 401}};
\node[box, below=0.85cm of screen] (full) {Studies retained for full-text eligibility assessment\\\textbf{n = 247} (235 eligible; 12 to resolve)};
\node[sidebox] (excl) at ($(screen)!0.5!(full) + (3.55cm,0)$)
  {Excluded after screening \textbf{n = 154}\\{E3:96, E5:23, E1:11, I6:10, E4:8, E2:6}};
\draw[arrow] (id) -- (dup);
\draw[arrow] (dup) -- (screen);
\draw[arrow] (screen) -- (full);
\draw[arrow] ($(screen)!0.5!(full)$) -- (excl.west);
\end{tikzpicture}
\caption{Current review flow in PRISMA-style summary form (updated corpus including Web of Science and the 2015--2026 time filter).}
\label{fig:prismaflow}
\end{figure}
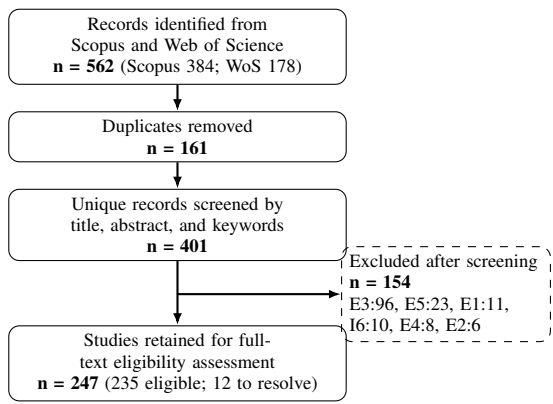

Fig.~\ref{fig:prismaflow} summarizes the current selection status. From 562 identified records (384 from Scopus and 178 from Web of Science), 161 duplicates were removed. The remaining 401 unique records were screened by title, abstract, and keywords. At this stage, 247 studies were retained for full-text eligibility assessment and 154 were excluded. Exclusions were dominated by records where coverage is not an objective, constraint, or evaluation metric (E3, 96 records), followed by surveys or conceptual works without their own evaluation (E5, 23), studies limited to low-level control, hardware, or sensing (E1, 11), records outside the 2015--2026 window (I6, 10), non-UAV platforms (E4, 8), and works using the UAV as a mere capture platform (E2, 6).
A preliminary keyword-assisted scan of the 247 retained studies reveals several trends. Coverage-oriented formulations dominate the set: 196 retained studies (about 79\%) explicitly mention area coverage, coverage path planning, or coverage maximization. Multi-UAV or cooperative settings appear in 118 studies (about 48\%), indicating that coordination and task distribution are central concerns. Energy-related constraints appear in 96 studies (about 39\%) and geometry-related concepts in 92 studies (about 37\%), whereas obstacle handling appears in 58 studies. In contrast, only 44 retained studies (about 18\%) explicitly mention weather, uncertainty, or dynamic environments.

The retained set is also strongly simulation-oriented: 147 studies mention simulation or simulated scenarios, whereas 46 mention field experiments, real-world environments, or real data. In methodological terms, the abstracts show a comparable presence of heuristic/constructive (63) and metaheuristic (63) cues, a growing presence of learning-based strategies (42), and a smaller set of exact-optimization cues (19). Regarding recency, 152 of the 247 retained studies (about 62\%) were published between 2023 and 2026, confirming a strong concentration of recent activity. Table~\ref{tab:signals} summarizes these non-exclusive signals.
\begin{table}[t]
\caption{Preliminary signals from retained studies ($n=247$)}
\label{tab:signals}
\centering\small
\begin{tabular}{p{0.67\columnwidth}r}
\toprule
Signal from title/abstract/keywords & Studies \\
\midrule
Coverage-oriented formulations & 196/247 \\
Multi-UAV or cooperative setting & 118/247 \\
Exact-optimization cues & 19/247 \\
Heuristic / constructive-pattern cues & 63/247 \\
Metaheuristic cues & 63/247 \\
Learning-based cues & 42/247 \\
Energy-aware constraints & 96/247 \\
Obstacle-related modeling & 58/247 \\
Geometry-related modeling & 92/247 \\
Weather/uncertainty/dynamics & 44/247 \\
Simulation-based validation & 147/247 \\
Real-world / field validation & 46/247 \\
\bottomrule
\end{tabular}
\end{table}
These results suggest that the literature already provides a substantial base for comparing coverage efficiency and energy-aware planning, but richer environmental realism remains less consistently modeled. This observation is relevant for environmental monitoring, where terrain, obstacles, sensing geometry, and uncertainty can critically alter route feasibility and information value. The preliminary evidence also hints that simulation remains the default validation strategy, which may limit external validity when methods are transferred to real missions.
\section{Threats to Validity}
\label{sec:threats}
At the current stage, four threats deserve explicit attention. First, the preliminary patterns in Table~\ref{tab:signals} were inferred from titles, abstracts, and keywords rather than from full-text coding, so they should be interpreted as screening-level lexical signals instead of final evidence; a term may appear (e.g., ``weather'') only as a future-work mention without being modeled. The full-text stage will replace these signals with verified, full-text coding. Second, screening up to this stage was performed by a single reviewer, which introduces potential selection and subjectivity bias. This was mitigated by applying explicit \emph{a priori} inclusion/exclusion criteria, documenting a decision and a criterion for every record, and adopting an objective-function rule to disambiguate the heterogeneous use of ``coverage''; borderline records were flagged and discussed with the second author. To strengthen reliability, the full-text eligibility stage will be double-coded by two reviewers and inter-rater agreement (Cohen's $\kappa$) will be reported. Third, restricting the sources to Scopus and Web of Science may miss venues better indexed elsewhere; this is mitigated by the broad overlap of both databases with IEEE Xplore and ACM content, and a complementary search in those interfaces is planned. Fourth, publication bias may favor studies with positive performance claims or cleaner simulation results. Terminology heterogeneity (coverage path planning, area coverage, exploration, informative planning) is additionally mitigated by the objective-function-based eligibility rule.
\section{Discussion and Ongoing Work}
The preliminary evidence indicates a field that is both active and fragmented. On one hand, the dominance of coverage formulations, multi-UAV coordination, and energy-aware planning suggests a maturing core agenda. On the other hand, the lower frequency of explicit weather, uncertainty, and obstacle-rich modeling indicates that many studies still validate under simplified conditions. This gap matters because environmental monitoring missions often operate over irregular geography, changing atmospheric conditions, and partially known spaces.

The retained corpus spans the expected algorithmic families: exact formulations such as MILP-based coverage routing \cite{cooperativeMILPCoverage2022,milpCoordinatedCoverage2023}; constructive heuristics and classic sweep or visibility patterns \cite{sweepCoverageMultiUAV2020,visibilityHeuristicVPP2020}; metaheuristics including genetic, swarm, and bio-inspired search, sometimes hybridized with constructive rules \cite{gaDisconnectedPolygons2024,terrain3DMetaheuristic2025,bwoPlantProtection2025,greedyACOCoverage2022}; geometry-driven decomposition over convex and concave regions \cite{forestConcaveCPP2025,tinyMLConvexCPP2024,capacityConstrainedCoverage2022}; informative path planning for environmental fields \cite{informativeTerrainIPP2020,pollutionSamplingIPP2024,continuousNonConvexIPP2026}; and learning-based coordination \cite{drlMultiUAVCoverage2025,marlLightweightCoverage2026,multiagentQLearning2021}. Energy-aware coverage planning and assignment remain central \cite{energyConstrainedCoverage2021,terrainEnergyCoverage2025,uavHumanDualLayer2025}, while learning and multi-agent coordination are increasingly visible in recent publications \cite{drlMultiUAVCoverage2025,improvedDQNCoverage2025,distributedCPPFramework2026}. This suggests that the field is moving toward richer coordination and adaptation mechanisms, but without a clear consensus yet on the most robust evaluation protocol or the most realistic combination of constraints.

The next stage of the review will therefore focus on full-text eligibility assessment, methodological quality scoring, and structured data extraction. The objective is not only to identify high-performing approaches, but also to clarify which combinations of optimization strategy, geometric representation, and operational constraints are most credible for realistic environmental monitoring scenarios. Those findings will directly inform the subsequent design of a coverage-maximization model grounded in computational geometry and realistic mission constraints.
\section{Conclusion}
This paper presented the protocol and preliminary screening results of an ongoing systematic literature review on autonomous UAV route planning for coverage maximization in environmental monitoring. The current evidence base already shows strong interest in coverage efficiency, coordination, and energy-aware planning, but suggests that uncertainty-rich and geometry-constrained scenarios remain comparatively underexplored. Completing the eligibility, quality, and extraction stages will enable a more rigorous synthesis of algorithmic trends, evaluation metrics, and open research gaps. These findings will guide the design of future UAV coverage-planning models integrating computational geometry and intelligent optimization techniques.
\bibliographystyle{IEEEtran}
\bibliography{CLEI2026WIPselectedreferences}

@article{prisma2021,
  title   = {The {PRISMA} 2020 statement: an updated guideline for reporting systematic reviews},
  author  = {Page, Matthew J. and McKenzie, Joanne E. and Bossuyt, Patrick M. and others},
  journal = {BMJ},
  volume  = {372},
  pages   = {n71},
  year    = {2021}
}

@article{energyConstrainedCoverage2021,
  author  = {Jensen-Nau, KR and Hermans, T and Leang, KK},
  title   = {Near-Optimal Area-Coverage Path Planning of Energy-Constrained Aerial Robots With Application in Autonomous Environmental Monitoring},
  journal = {IEEE Trans. Autom. Sci. Eng.},
  volume  = {18},
  number  = {3},
  pages   = {1453--1468},
  year    = {2021},
  doi     = {10.1109/TASE.2020.3016276},
}

@article{terrainEnergyCoverage2025,
  author  = {Shao, Q and Mao, XF and Xu, WB},
  title   = {Energy-Aware UAV Coverage Planning in Mountainous Terrain via Contour-Aligned Path Generation},
  journal = {IEEE Robot. Autom. Lett.},
  volume  = {10},
  number  = {12},
  pages   = {12373--12380},
  year    = {2025},
  doi     = {10.1109/LRA.2025.3621932},
}

@article{forestConcaveCPP2025,
  author  = {Kang, B and Wang, CA and Su, Y and Zeng, JY},
  title   = {Multi-UAV forest area inspection path planning based on concave polygon region decomposition},
  journal = {Sci. Rep.},
  volume  = {15},
  number  = {1},
  year    = {2025},
  doi     = {10.1038/s41598-025-26060-7},
}

@INPROCEEDINGS{drlMultiUAVCoverage2025,
  author={Liu, Qian and Zhang, Yong and Jia, Lu},
  booktitle={2025 2nd International Conference on Machine Learning, Pattern Recognition and Automation Engineering (MLPRAE)}, 
  title={A Deep Reinforcement Learning Approach for Multi-UAV Collaborative Coverage with Adaptive Step Size and Dynamic Reward Mechanism}, 
  year={2025},
  volume={},
  number={},
  pages={7-10},
  doi={10.1109/MLPRAE67267.2025.11290942}}

@article{sweepCoverageMultiUAV2020,
  author  = {Li, J and Xiong, YH and She, JH and Wu, M},
  title   = {A Path Planning Method for Sweep Coverage With Multiple UAVs},
  journal = {IEEE Internet Things J.},
  volume  = {7},
  number  = {9},
  pages   = {8967--8978},
  year    = {2020},
  doi     = {10.1109/JIOT.2020.2999083},
}

@article{informativeTerrainIPP2020,
  author  = {Popovic, M and Vidal-Calleja, T and Hitz, G and Chung, JJ and Sa, I and Siegwart, R and Nieto, J},
  title   = {An informative path planning framework for UAV-based terrain monitoring},
  journal = {Auton. Robots},
  volume  = {44},
  number  = {6},
  pages   = {889--911},
  year    = {2020},
  doi     = {10.1007/s10514-020-09903-2},
}

@INPROCEEDINGS{cooperativeMILPCoverage2022,
  author={Zhang, Furong and Zhang, Xiaopan},
  booktitle={2022 9th International Conference on Dependable Systems and Their Applications (DSA)}, 
  title={Cooperative Area Coverage Path Planning for Multiple UAVs Over Large Areas}, 
  year={2022},
  volume={},
  number={},
  pages={346-352},
  doi={10.1109/DSA56465.2022.00053}}

@article{milpCoordinatedCoverage2023,
  author  = {Zhang, XP and Zhang, FR and Tang, Z and Chen, XJ},
  title   = {A MILP model on coordinated coverage path planning system for UAV-ship hybrid team scheduling software},
  journal = {J. Syst. Softw.},
  volume  = {206},
  year    = {2023},
  doi     = {10.1016/j.jss.2023.111854},
}

@article{visibilityHeuristicVPP2020,
  author  = {Sanchez-Fernandez, AJ and Romero, LF and Bandera, G and Tabik, S},
  title   = {VPP: Visibility-Based Path Planning Heuristic for Monitoring Large Regions of Complex Terrain Using a UAV Onboard Camera},
  journal = {IEEE J. Sel. Topics Appl. Earth Observ. Remote Sens.},
  volume  = {15},
  pages   = {944--955},
  year    = {2022},
  doi     = {10.1109/JSTARS.2021.3134948},
}

@article{greedyACOCoverage2022,
  author  = {Jia, YH and Zhou, SB and Zeng, Q and Li, CQ and Chen, D and Zhang, KZ and Liu, LY and Chen, ZY},
  title   = {The UAV Path Coverage Algorithm Based on the Greedy Strategy and Ant Colony Optimization},
  journal = {Electronics},
  volume  = {11},
  number  = {17},
  year    = {2022},
  doi     = {10.3390/electronics11172667},
}

@INPROCEEDINGS{terrain3DMetaheuristic2025,
  author={Ouyang, Nan and Xie, Jiahui and Lin, Feng},
  booktitle={2025 6th International Conference on Computer Engineering and Application (ICCEA)}, 
  title={Low-Altitude UAV Trajectory Optimization for Complex 3D Terrains Based on Energy Consumption}, 
  year={2025},
  volume={},
  number={},
  pages={01-08},
  doi={10.1109/ICCEA65460.2025.11103265}}

@InProceedings{gaDisconnectedPolygons2024,
author="Hu, Tianmi
and Wang, Shuyue
and Lyu, Yang
and Liang, Xinkai
and Pan, Quan",
title="Coverage Path Planning of Multiple Disconnected Convex Polygons Based on Improved Genetic Algorithm",
booktitle="Proceedings of 2024 12th China Conference on Command and Control",
year="2024",
publisher="Springer Nature Singapore",
address="Singapore",
pages="55--67",
isbn="978-981-97-7774-7"
}

@article{bwoPlantProtection2025,
  author  = {Tan, ZP and Huang, KC and Tang, Y and Fang, MW and Huang, HS},
  title   = {Multi-area coverage path planning for plant protection UAVs based on a hybrid strategy beluga whale optimization algorithm},
  journal = {Smart Agric. Technol.},
  volume  = {12},
  year    = {2025},
  doi     = {10.1016/j.atech.2025.101379},
}

@article{marlLightweightCoverage2026,
  author  = {Qian, ZC and Feng, Y and Liu, NB and Qian, Q},
  title   = {CLMPO-EC: A Lightweight Multi-UAV Multiarea Coverage Path Planning Method Using Deep Reinforcement Learning},
  journal = {IEEE Internet Things J.},
  volume  = {13},
  number  = {8},
  pages   = {16535--16549},
  year    = {2026},
  doi     = {10.1109/JIOT.2026.3659864},
}

@article{multiagentQLearning2021,
  author  = {Akin, E and Demir, K and Yetgin, H},
  title   = {Multiagent Q-learning based UAV trajectory planning for effective situational awareness},
  journal = {Turk. J. Electr. Eng. Comput. Sci.},
  volume  = {29},
  number  = {5},
  pages   = {2561--2579},
  year    = {2021},
  doi     = {10.3906/elk-2012-41},
}

@article{tinyMLConvexCPP2024,
  author  = {Jia, B and Gao, Z and Jing, JQ and Huang, BQ and Liu, S and Muhammad, K and Rodrigues, JJPC},
  title   = {Coverage Path Planning for IoUAVs With Tiny Machine Learning in Complex Areas Based on Convex Decomposition},
  journal = {IEEE Internet Things J.},
  volume  = {11},
  number  = {12},
  pages   = {21103--21111},
  year    = {2024},
  doi     = {10.1109/JIOT.2024.3361857},
}

@ARTICLE{capacityConstrainedCoverage2022,
  author={Agarwal, Saurav and Akella, Srinivas},
  journal={IEEE Robot. Autom. Lett.}, 
  title={Area Coverage With Multiple Capacity-Constrained Robots}, 
  year={2022},
  volume={7},
  number={2},
  pages={3734-3741},
  doi={10.1109/LRA.2022.3146952}}

@article{pollutionSamplingIPP2024,
  author  = {Kosior, M and Przystalka, P and Panfil, W},
  title   = {Adaptive Path Planning for UAV-Based Pollution Sampling},
  journal = {Appl. Sci.},
  volume  = {14},
  number  = {24},
  year    = {2024},
  doi     = {10.3390/app142412065},
}

@article{continuousNonConvexIPP2026,
  author  = {Hao, H and Silvestre, D and Silvestre, C},
  title   = {Continuous trajectory planning for non-convex utility functions using hybrid optimization},
  journal = {Eur. J. Control},
  volume  = {87},
  year    = {2026},
  doi     = {10.1016/j.ejcon.2025.101425},
}

@article{distributedCPPFramework2026,
  author  = {Adoni, WYH and Lorenz, S and Gloaguen, R and Singh, A and Kühne, TD},
  title   = {A distributed coverage path planning framework for autonomous unmanned aerial vehicle (UAV) swarms},
  journal = {Expert Syst. Appl.},
  volume  = {322},
  year    = {2026},
  doi     = {10.1016/j.eswa.2026.132382},
}

@article{improvedDQNCoverage2025,
  author  = {Ni, JJ and Ge, YC and Zhao, YH and Gu, Y},
  title   = {An Improved Multi-UAV Area Coverage Path Planning Approach Based on Deep Q-Networks},
  journal = {Appl. Sci.},
  volume  = {15},
  number  = {20},
  year    = {2025},
  doi     = {10.3390/app152011211},
}

@article{uavHumanDualLayer2025,
  author  = {Yang, G and Mo, YD and Lv, CY and Zhang, Y and Li, J and Wei, SM},
  title   = {A dual-layer task planning algorithm based on UAVs-human cooperation for search and rescue},
  journal = {Appl. Soft Comput.},
  volume  = {181},
  year    = {2025},
  doi     = {10.1016/j.asoc.2025.113488},
}
\end{document}